%% file: camera-ready_GLSFormer.tex
\documentclass[runningheads]{llncs}
\usepackage{hyperref}
\usepackage{graphicx}
\input{commands}
\usepackage{amsmath}
\usepackage{tabularx}
\usepackage{booktabs}
\usepackage{multirow}
\usepackage{soul}
\usepackage{ulem}

\usepackage[misc]{ifsym}

\begin{document}
\title{\GLSF: Gated - Long, Short Sequence Transformer for Step Recognition in Surgical Videos}
\author{Nisarg A. Shah\inst{1,}\Letter \and 
Shameema Sikder\inst{2,3} \and 
S. Swaroop Vedula\inst{3} \and 
Vishal M. Patel\inst{1} 
}
\authorrunning{N. Shah et al.}
\titlerunning{\GLSF : Gated - Long, Short Sequence Transformer for Step Recognition}
\institute{Johns Hopkins University, Baltimore, MD 21218, USA \and
Wilmer Eye Institute, Johns Hopkins University School of Medicine, Baltimore, MD \and 
Malone Center for Engineering in Healthcare, Johns Hopkins University
\\\email{snisarg812@gmail.com}}
\maketitle              
\begin{abstract}
Automated surgical step recognition is an important task that can significantly improve patient safety and decision-making during surgeries. Existing state-of-the-art methods for surgical step recognition either rely on separate, multi-stage modeling of spatial and temporal information or operate on short-range temporal resolution when learned jointly. However, the benefits of joint modeling of spatio-temporal features and long-range information are not taken in account. In this paper, we propose a vision transformer-based approach to jointly learn spatio-temporal features directly from sequence of frame-level patches. Our method incorporates a gated-temporal attention mechanism that intelligently combines short-term and long-term spatio-temporal feature representations. We extensively evaluate our approach on two cataract surgery video datasets, namely Cataract-101 and D99, and demonstrate superior performance compared to various state-of-the-art methods. These results validate the suitability of our proposed approach for automated surgical step recognition. Our code is released at: \href{https://github.com/nisargshah1999/GLSFormer}{https://github.com/nisargshah1999/GLSFormer}

\end{abstract}
\section{Introduction}
Surgical step recognition is necessary to enable downstream applications such as surgical workflow analysis \cite{bricon2007context,yu2019assessment}, context-aware decision support \cite{padoy2019machine}, anomaly detection \cite{huaulme2020offline}, and record-keeping purposes \cite{zisimopoulos2018deepstep,czempiel2020tecno}. 
Some factors that make recognition of steps in surgical videos a challenging problem \cite{lecuyer2020assisted,padoy2019machine} include variability in patient anatomy and surgeon style \cite{funke2019video}, similarities across steps in a procedure \cite{czempiel2020tecno,jin2021temporal}, online recognition\cite{yi2019hard} and scene blur \cite{padoy2019machine,gao2021trans}.

Early statistical methods to recognize surgical workflow, such as Conditional Random Fields \mbox{\cite{lea2015improved,tao2013surgical}}, Hidden Markov Models (HMMs) \mbox{\cite{dergachyova2016automatic,twinanda2016endonet}} and Dynamic Time Warping \mbox{\cite{lalys2013automatic,blum2010modeling}}, have limited representation capacity due to pre-defined dependencies.
Multiple deep learning based methods have also been proposed for surgical step recognition. SV-RCNet \cite{jin2017sv} jointly trains ResNet \cite{he2016deep} and a long short-term memory (LSTM) model and uses a prior knowledge scheme during inference. TMRNet \cite{jin2021temporal} utilizes a memory bank to store long range information on the relationship between the current frame and its previous frames. SV-RCNet and TMRNet use LSTMs, which are constrained in capturing long-term dependencies in surgical videos due to their limited temporal memory. Furthermore, LSTMs process information in a sequential manner that results in longer inference times. Methods that don't use LSTMs include a 3D-covolutional neural network (3D-CNN) to learn spatio-temporal features \cite{funke2019using} and temporal convolution networks (TCNs) \cite{zhang2020symmetric}. In recent work, a two-stage network called TeCNO included a ResNet to learn spatial features, which are then modeled with a multi-scale TCN to capture long-term dependencies \cite{czempiel2020tecno}. The previous networks use multi-stage training to exploit spatial and temporal information separately. This approach limits model capacity to learn spatial features using the temporal information. Furthermore, the temporal modeling does not sufficiently benefit from low-dimensional spatial features resulting in low sensitivity to identify transitions between activities \cite{jin2017sv,gao2021trans}. \cite{gao2021trans} attempts to address this issue by adding one more stage to the network using a feature fusion technique that employs a transformer to refine features extracted from the first and second stages of TeCNO \cite{czempiel2020tecno}.

Transformers facilitate improved feature representation by effectively modeling long-term dependencies, which is important for recognizing surgical steps in complex videos. In addition, they offer faster processing speeds due to their parallel computing architecture. To exploit these benefits and address the issue of inductive bias (e.g. local connectivity and translation equivariance) in CNNs, recent methods use only transformers as their building blocks, i.e., Vision Transformer. For example, TimesFormer \cite{bertasius2021space}, applies self-attention mechanisms to learn the spatial and temporal relations in videos, and ViViT \cite{arnab2021vivit} utilizes a vision transformer architecture \cite{dosovitskiy2020image} for video recognition. However, these models were proposed for temporal clips and do not specifically focus on capturing long-range dependencies, which is important for surgical step recognition in long-duration untrimmed videos.

In this work, we propose a transformer-based model with the following contributions: (1) Spatio-temporal attention is used as the building blocks to address the issues of inductive bias and end-to-end learning of surgical steps; 
(2) A two-stream model, called \textbf{G}ated - \textbf{L}ong, \textbf{S}hort sequence Trans\textbf{former} \GLSF, is proposed to capture long-range dependencies and a gating module to leverage cross-stream information in its latent space; and (3) We show the effectiveness of the proposed approach by extensively evaluating GLSFormer on two cataract surgery video datasets and show that it outperforms all compared methods. 
\vspace{-2mm}
\section{The \GLSF~model}
\vspace{-2mm}
Given an untrimmed video with $X[1:T]$ frames, where $T$ represents the total number of frames in the video, the objective is to predict the step $Y[t]$ of a given frame at time $t$. 
\\
\begin{figure}[h]
\centering
  \includegraphics[page=5,width=1\linewidth]{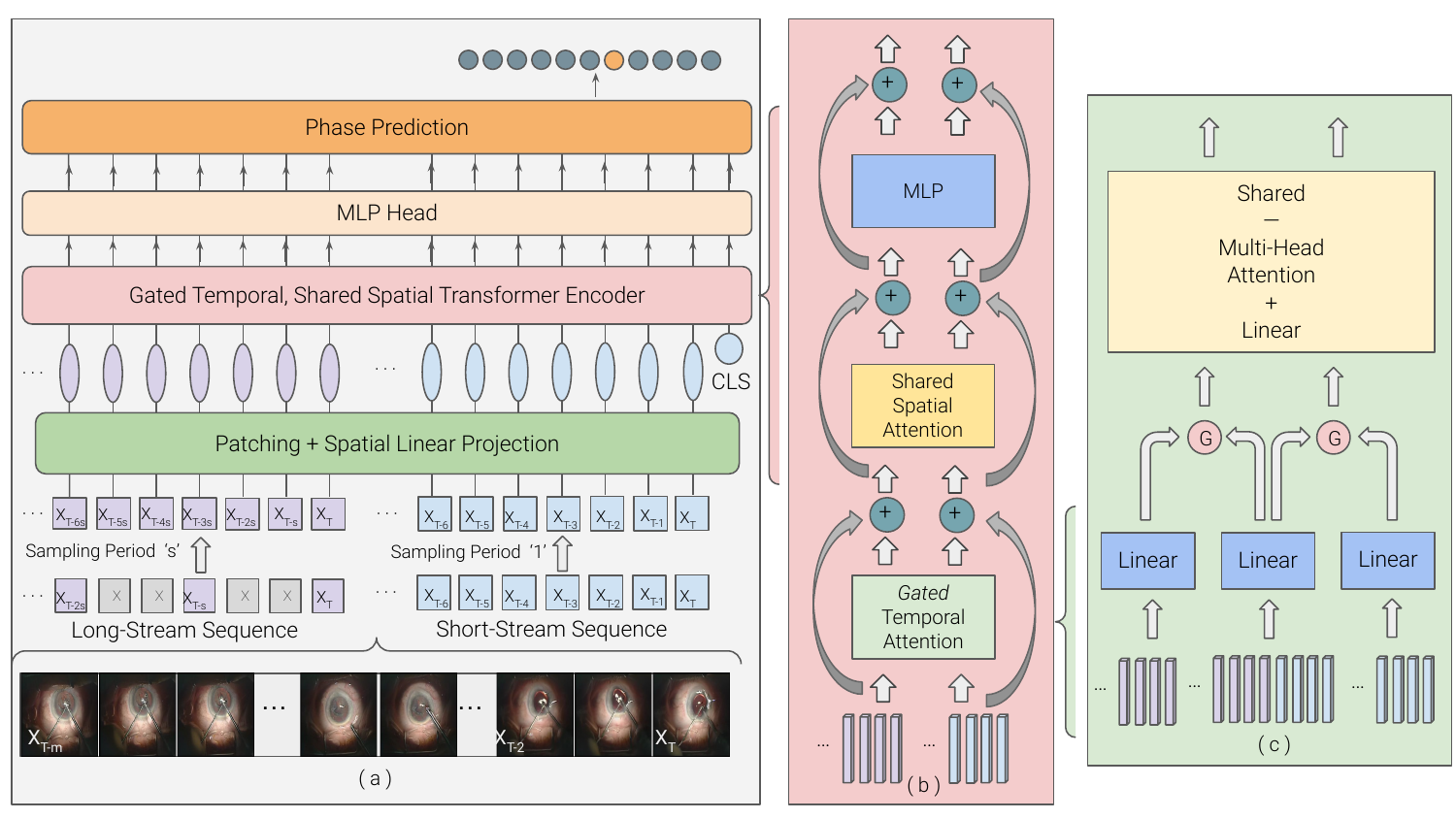}
 \vskip-10pt \caption{Overview of the proposed \GLSF. Specifically, \GLSF takes two streams (long-stream and short-stream) image sequences as input, sampled with sampling period of $s$ and $1$ respectively. Later, each frame is decomposed into non-overlapping patches and each of these patches are linearly mapped to an embedding vector.
 These embedded features are spatio-temporally encoded into a feature representation using a sequential temporal-spatial attention block as shown in (b). Architecture of gated-temporal attention is showed in detail in (c). The final feature representation is then examined by a multilayer perceptron head and a linear layer to produce a step prediction at every time-point. Our method provides a single-stage, end-to-end trainable model for surgical step recognition.}
  \label{fig:system-overview}
\end{figure}
\noindent {\textbf{Long-short Sequence.}}
We propose \GLSF model that can capture both short-term and long-term dependencies in surgical videos. The input to our \GLSF are two video sequences, a short-term sequence consisting of the last $n_{st}$ frames from time $t$, and a long-term sequence composed of $n_{lt}$ frames selected from a sub-sampled set of frames with a sampling period of $s$.
The long-term sequence provides a coarse overview of information distant in time and can aid in accurate predictions of the current frame, overcoming false prediction based on common artifacts in short-term sequences. In addition, the overview of information can address the high variability in surgical scenes \cite{gao2021trans,czempiel2020tecno,jin2021temporal,feichtenhofer2019slowfast}. By leveraging both short-term and long-term sequences, our model can accurately capture the complex context present in long surgical videos.\\
\noindent {\bf{Patch Embedding.}}  
We decompose each frame of dimension $H\times W\times 3$ into $N$ non-overlapping patches of size $Q \times Q$ where $N = \frac{HW}{Q^2}$. Each patch is flattened into a vector $x_{p,t} \in \mathbb{R}^{3Q^2}$ for each frame $t$ and spatial location, $p \in (1,N) $. We linearly map the patches of short and long term videos frames into embedding vector  of dimension $\mathbb{R}^{K}$ using a shared learnable matrix $E \in \mathbb{R}^{K \times 3Q^2}$.  We concatenate the patch embeddings of the short and long-term streams along the frame dimension to form feature representations $x_{p,t}^{st}$ and $x_{p,t}^{lt}$ of size $N_{} \times n_{st} \times K$ and $N_{} \times n_{lt} \times K$, respectively. along with a learnable positional embedding $e_{p,t}^{st-pos}$ and $e_{p,t}^{lt-pos}$ to encode spatio-temporal position as follows 
\setlength{\belowdisplayskip}{0pt} \setlength{\belowdisplayshortskip}{0pt}
\setlength{\abovedisplayskip}{0pt} \setlength{\abovedisplayshortskip}{0pt}
\begin{equation}
z_{p,t}^{st} = Ex_{p,t}^{st} + e_{p,t}^{st-pos},\;\;\;\;\;
z_{p,t}^{lt} = Ex_{p,t}^{lt} + e_{p,t}^{lt-pos}.
\label{eqn:pos}
\end{equation}
Note that a special learnable vector $z_{0,0}^{st} \in \mathbb{R}^K$ representing the step classification token is added in the first position. Our approach is similar to word embeddings in NLP transformer models~\cite{devlin2018bert,bertasius2021space,arnab2021vivit}. \\
\noindent {\bf{Gated Temporal, Shared Spatial Transformer Encoder.}}
Our Transformer Encoder, consisting of Gated Temporal Attention module and Shared Spatial Attention module takes the sequence of embedding vectors $z_{p,t}^{st}$ and $z_{p,t}^{lt}$ as input. 
In a self-attention module for spatio-termporal models, computational complexity increases non-linearly O($T^{2}S^{2}$) with increase in spatial resolution(S) or temporal frames(T). Thus, to reduce the complexity, we sequentially process our gated temporal cross attention module and spatial attention module \cite{bertasius2021space,arnab2021vivit}. 
Transformer Encoder consists of L Gated-Temporal, Spatial Attention blocks. At each block $l$, feature representation is computed for both streams from the representation $z_{l-1}^{lt}$ and $z_{l-1}^{st}$ encoded by the preceding block $l-1$.
We explain our Gated Temporal attention and shared spatial attention in more detail in the rest of the section\\

\noindent {\textbf{\textit{Gated Temporal Attention.}}} 
The temporal cross-attention module aligns the long-term($z_{l-1}^{lt}$) and short-term features($z_{l-1}^{st}$) in the temporal domain, allowing the model to better capture the relationship between the long and short term streams. We concatenate both of these streams to form a strong joint stream that has both fine-grained information from the short-term stream and global context information from the long-term stream. 
Firstly, a query/key/value vector for each patch in the representations $z_{l-1}^{lt}(p,t)$, $z_{l-1}^{st}(p,t)$ and $z_{l-1}^{lt,st}(p,t)$ using linear transformations with weight matrices $U_{qkv}^{lt}$, $U_{qkv}^{st}$ and $U_{qkv}^{lt,st}$ respectively and normalization using LayerNorm is computed as follows:
\begin{equation}
\label{eqn:qkv_st}
\begin{aligned}
\relax
(QKV)^{a; st} &= z^{st} U_{qkv}^{st}, \
&\qquad \qquad ~ U_{qkv}^{st} &\in \mathbb{R}^{K_{h} \times 3K_{h}} \\
(QKV)^{a; lt} &= z^{lt} U_{qkv}^{lt}, \
&\qquad \qquad ~ U_{qkv}^{lt} &\in \mathbb{R}^{K_{h} \times 3K_{h}} \\
(QKV)^{a; lt,st} &= [z^{lt} \oplus z^{st}]  U_{qkv}^{lt,st}, \
&\qquad \qquad ~ U_{qkv}^{lt,st} &\in \mathbb{R}^{K_{h} \times 3K_{h}}
\end{aligned}
\end{equation}
 where $a$ ranges from 1 to $A$ representing attention heads. The total number of attention heads is denoted by $A$, and each has a latent dimensionality of $K_h = \frac{K}{A}$. The computation of these QKV vectors is essential for  multi-head attention in transformer.
 
Now for refining the streams, with most relevant cross-stream information, we 
gate the individual stream's temporal features$(I)$ $(QKV)^{a;lt/st}$ with the joint stream temporal features$(J)$ $(QKV)^{a;lt,st}$. 
Gating parameters are calculated by concatenating $I$ and $J$ and passing them through linear and softmax layers which  predict $Gt^{st}$ and $Gt^{lt}$ for $(QKV)^{a;st}$ and $(QKV)^{a;lt}$, respectively. By gating the individual stream's temporal features with the joint stream temporal features, the model is able to selectively attend to the most relevant features from both streams, resulting in a more informative representation. This helps in capturing complex relationships between the streams and improves the overall performance of the model.  This computation can be described as follows
\begin{equation}
\begin{aligned}
\relax
[q_{gt}^{st},k_{gt}^{st},v_{gt}^{st}] &= (1 - Gt^{st}(0))*(QKV)^{a;st} + Gt^{st}(0)*(QKV)^{a;lt,st}[lt:lt+st] \\
\nonumber [q_{gt}^{lt},k_{gt}^{lt},v_{gt}^{lt}] &= (1 - Gt^{lt}(0))*(QKV)^{a;lt} + Gt^{lt}(0)*(QKV)^{a;lt,st}[:lt]. 
\end{aligned}
\end{equation} 

Later, temporal attention is computed by comparing each patch $(p, t)$ with all patches at the same spatial location in other frames of both streams, as follows 
\begin{equation}
\begin{aligned}
\nonumber \alpha_{gt,(p,t)}^{\text{($\bullet$)-temporal}(l,a)} &= \text{softmax}\begin{pmatrix}\frac{q_{gt,(p,t)}^{(\bullet),(l,a)}}{\sqrt{K_{h}}} 
\left[k_{gt,(0,0)}^{(l,a)} \right. \left. \prod_{t'=1}^{n_{lt}} k_{gt,(p,t')}^{lt,(l,a)} \prod_{t'=1}^{n_{st}} k_{gt,(p,t')}^{st,(l,a)} \right] \end{pmatrix}
\end{aligned}
\label{eqn:temp_att}
\end{equation}
Here, $\alpha_{gt,(p,t)}^{\text{($\bullet$)-temporal}(l,a)}$ is separately calculated for the long-stream and short-stream, where ($\bullet$) = lt or st.  Similar to the vision transformer, encoding blocks for each layer ($z^{lt}$ and $z^{st}$) are computed by taking the weighted sum of value vectors ($\text{SA}_{a}(z)$) using self-attention coefficients from each attention head as follows
\begin{equation}
\text{SA}_{a}(z = z^{lt} \oplus z^{st}) = \large{(\alpha^{lt,a}_{gt} \oplus \alpha^{st,a}_{gt}) \cdot (v^{lt,a}_{gt} \oplus v^{st,a}_{gt})}.
\end{equation}
Next, the self-attention block ($\text{SA}_{a}(z)$) for each attention head is projected along with a residual connection from the previous layer. This multi-head self-attention (MSA) operation can be described as follows
\begin{equation}
\begin{aligned}
\relax
\text{MSA}(z)=[\text{SA}_1(z); \text{SA}_2(z); ... ; \text{SA}_A(z)] \times U{msa}, \quad \quad U{msa} \in \mathbb{R}^{k \cdot D_h \times D} \
\end{aligned}
\label{eqn:msa}
\end{equation}
\vspace{-3mm}
\begin{equation}
\begin{aligned}
\relax
(z^{'})^{l} &= \text{MSA}(z^{l}) + (z)^{l-1}
\end{aligned}
\label{eqn:residual}
\end{equation}
Here, $(z^{'})^{l}$ is the concatenation of $z^{lt}$ and $z^{st}$. 

\noindent {\textbf{\textit{Shared Spatial Attention.}}} 
Next, we apply self-attention to the patches of the same frame to capture spatial relationships and dependencies within the frame. To accomplish this, we calculate new key/query/value using Eqn.~\eqref{eqn:qkv_st} and use it to perform spatial attention in Eqn.~\eqref{eqn:spat_att}. 
\begin{equation}
\begin{aligned}
\alpha_{gt,(p,t)}^{\text{($\bullet$)-spatial}(l,a)} &= \text{softmax}\begin{pmatrix}\frac{q_{gt,(p,t)}^{(\bullet),(l,a)}}{\sqrt{K_{h}}} 
\left[k_{gt,(0,0)}^{(l,a)} \right. \left. \prod_{p'=1}^{N} k_{gt,(p',t)}^{(\bullet),(l,a)} \right] \end{pmatrix}
\end{aligned}
\label{eqn:spat_att}
\end{equation}
The encoding blocks are also calculated using Eqn.~\eqref{eqn:msa} and \eqref{eqn:residual}, and the resulting vector is passed through a multilayer perceptron (MLP) of Eqn.~\eqref{eqn:mlp} to obtain the final encoding $z^{\prime}(p,t)$ for the patch at block $l$ as follows
\begin{equation}
\begin{aligned}
\relax
(z)^{l} &= \text{MLP}(LN((z^{'})^{l})) + (z^{'})^{l},\;\;\;[z^{lt},z^{st}] &= z^{l}. \\
\end{aligned}
\label{eqn:mlp}
\end{equation}
The embedding for the entire clip is obtained by taking the output from the final block and passing it through a MLP with one hidden layer.  The corresponding computation can be described as  $y = \text{LN}(\text{z}^{L}_{(0,0)}) \in \mathbb{R}^D.$ 
The classification token is used as the final input to the MLP for predicting the step class at time $t$. Our \GLSF is trained using the cross-entropy loss.
\vspace{-2mm}
\section{Experiments and Results}
\vspace{-2mm}
\noindent {\bf{Datasets.}} We evaluate our GLS-Former on two video datasets of cataract surgery, namely Cataract-101~\cite{schoeffmann2018cataract} and D99~\cite{yu2019assessment}. The Cataract-101 dataset comprises of 101 cataract surgery video recordings, each captured at 25 frames per second and annotated into 10 steps by surgeons. The spatial resolution of these videos is $720\times540$, and temporal resolution of 25 \textit{fps}. In accordance with previous studies~\cite{gao2021trans,twinanda2016endonet}, we use 50 videos for training, 10 for validation and 40 videos for testing. The D99 dataset, which consists of 99 videos with temporal segment annotations of 12 steps by expert physicians, has a frame resolution of $640\times480$ at 59 \textit{fps}. We randomly shuffled videos and select 60, 20 and 19 videos for training, validation and testing respectively. All videos are subsampled to 1 frame per second, as done in previous studies \cite{gao2021trans,twinanda2016endonet}, and the frames are resized to $250\times250$ resolution.\\
\noindent {\bf{Evaluation Metrics.}}
To accurately evaluate the results of surgical step prediction models, we use four different metrics, namely Accuracy, Precision, Recall, and Jaccard index \cite{gao2021trans,czempiel2020tecno,twinanda2016endonet}.\\ 
\noindent {\bf{Implementation Details.}}
We utilized an NVIDIA RTX A5000 GPU to train our \GLSF model on PyTorch. The batch size was set equal to 64. Data augmentations were applied  including $224\times224$ cropping, random mirroring, and color jittering. We employed the Adam optimizer with an initial learning rate of 5e-5 for 50 epochs. Additionally, we initialized the shared parameters of the model from a pre-trained model on Kinetics-400 \cite{kay2017kinetics,bertasius2021space}. The model's depth and the number of attention heads were set equal to 12 each. We used 8 frames for both short-stream and long-stream, and sampling rate of 8, unless stated otherwise.\\
\begin{table*}[h]
\centering
\footnotesize
\vspace{-7mm}
\caption{Quantitative results of step recognition from different methods on the Cataract-101 and D99 datasets. The average metrics over five repetitions of the experiment with different data partitions for cross-validation are reported (\%) along with their respective standard deviation ($\pm$).}
\resizebox{\textwidth}{!}{%
\begin{tabular}{ccccccccc} \toprule
\multirow{2}{*}{Method} & \multicolumn{4}{c}{Cataract-101}       & \multicolumn{4}{c}{D99}               \\   \cline{2-5}\cline{5-9} 
            & Accuracy & Precision & Recall & Jaccard & Accuracy & Precision & Recall & Jaccard  \\ \toprule
ResNet\cite{he2016deep}     & 82.64 $\pm$ 1.54 & 76.68 $\pm$ 1.86 & 74.73 $\pm$ 1.27 & 62.58 $\pm$ 1.92 & 72.06 $\pm$ 2.12 & 54.76 $\pm$ 2.77 & 52.28 $\pm$ 2.89 & 37.98 $\pm$ 2.97        \\
SV-RCNet\cite{jin2017sv}    & 86.13 $\pm$ 0.91 & 84.96 $\pm$ 0.94 & 76.61 $\pm$ 1.18 & 66.51 $\pm$ 1.30 &  73.39 $\pm$ 1.64 & 58.18 $\pm$ 1.67 & 54.25 $\pm$ 1.86 & 39.15 $\pm$ 2.03     \\
OHFM\cite{yi2019hard}        & 87.82 $\pm$ 0.71 & 85.37 $\pm$ 0.78 & 78.29 $\pm$ 0.81 & 69.01 $\pm$ 0.93 &  73.82 $\pm$ 1.13 & 59.12 $\pm$ 1.33 & 55.49 $\pm$ 1.63 & 40.01 $\pm$ 1.68          \\
TeCNO\cite{czempiel2020tecno}       & 88.26 $\pm$ 0.92 & 86.03 $\pm$ 0.83 & 79.52 $\pm$ 0.90 & 70.18 $\pm$ 1.15 &  74.07 $\pm$ 1.78 & 61.56 $\pm$ 1.41 & 55.81 $\pm$ 1.58 & 41.31 $\pm$ 1.72        \\
TMRNet\cite{jin2021temporal} & 89.68 $\pm$ 0.76 & 85.09 $\pm$ 0.72 & 82.44 $\pm$ 0.75 & 71.83 $\pm$ 0.91 &  75.11 $\pm$ 0.91 & 61.37 $\pm$ 1.46 & 56.02 $\pm$ 1.65 & 41.42 $\pm$ 1.76      \\ 
Trans-SVNet\cite{gao2021trans} & 89.45 $\pm$ 0.88 & 86.72 $\pm$ 0.85 & 81.12 $\pm$ 0.93 & 72.32 $\pm$ 1.04 &  74.89 $\pm$ 1.37 & 60.12 $\pm$ 1.55 & 56.36 $\pm$ 1.24 & 42.06 $\pm$ 1.51       \\  \midrule
ViT\cite{dosovitskiy2020image}         & 84.56 $\pm$ 1.72 & 78.51 $\pm$ 1.42 & 75.62 $\pm$ 1.83 & 64.77 $\pm$ 1.97 &  72.45 $\pm$ 1.91 & 55.15 $\pm$ 2.42 & 53.60 $\pm$ 2.63 & 38.18 $\pm$ 2.79       \\
TimesFormer\cite{bertasius2021space} & 90.76 $\pm$ 1.05 & 85.38 $\pm$ 0.93 & 84.47 $\pm$ 0.95 & 75.97 $\pm$ 1.26 &   77.83 $\pm$ 0.96 & 64.24 $\pm$ 1.20 & 55.17 $\pm$ 1.26 & 42.69 $\pm$ 1.34       \\
GLSFormer   & \textbf{92.91 $\pm$ 0.67} & \textbf{90.04 $\pm$ 0.71} & \textbf{89.45 $\pm$ 0.79} & \textbf{81.89 $\pm$ 0.92} & \textbf{80.24 $\pm$ 1.02} & \textbf{69.98 $\pm$ 1.09} & \textbf{56.07 $\pm$ 1.12} & \textbf{48.35 $\pm$ 1.22}    \\ \bottomrule \\
\end{tabular}}
\label{tbl:table_2}
\end{table*}
\vspace{-8mm}

\noindent {\bf{Comparison with state-of-the-art methods.}}
Table~\ref{tbl:table_2} presents a comparison between our proposed approach, \GLSF, and current state-of-the-art methods for surgical step prediction. The comparison includes six models (1-6) that utilize ResNet~\cite{he2016deep} as a spatial feature extractor and two models (7-8) that use vision transformer backbones specifically designed for surgical step prediction. 
While SV-RCNet, OHFM, and TMRNet use ResNet-LSTM to capture short-range dependencies, OHFM uses a multi-step framework and TMRNet uses a multi-stage network to refine predictions using non-trainable long-range memory banks. Our approach achieves a significant improvement of 7\%-10\% in Jaccard index using a simpler, single-stage training procedure with higher temporal resolution. Although other multi-stage approaches like TeCNO and Trans-SVNet use temporal convolutions to capture long-range dependencies, we achieve a boost of 6\%-9\% with joint spatiotemporal modeling.
In contrast, transformer-based models capture spatial information (ViT) and short-term temporal (TimesFormer) efficiently, but they lack the long-term coarse step information required for complex videos. Our approach combines short-term and long-term spatiotemporal information in a single stage using gated-temporal and shared-spatial attention. This approach outperforms ViT and TimesFormer by a relative improvement of 6\%-11\% in Jaccard index across both datasets. \\

\begin{figure}[htp!]
\centering
  \includegraphics[page=2,width=.95\linewidth]{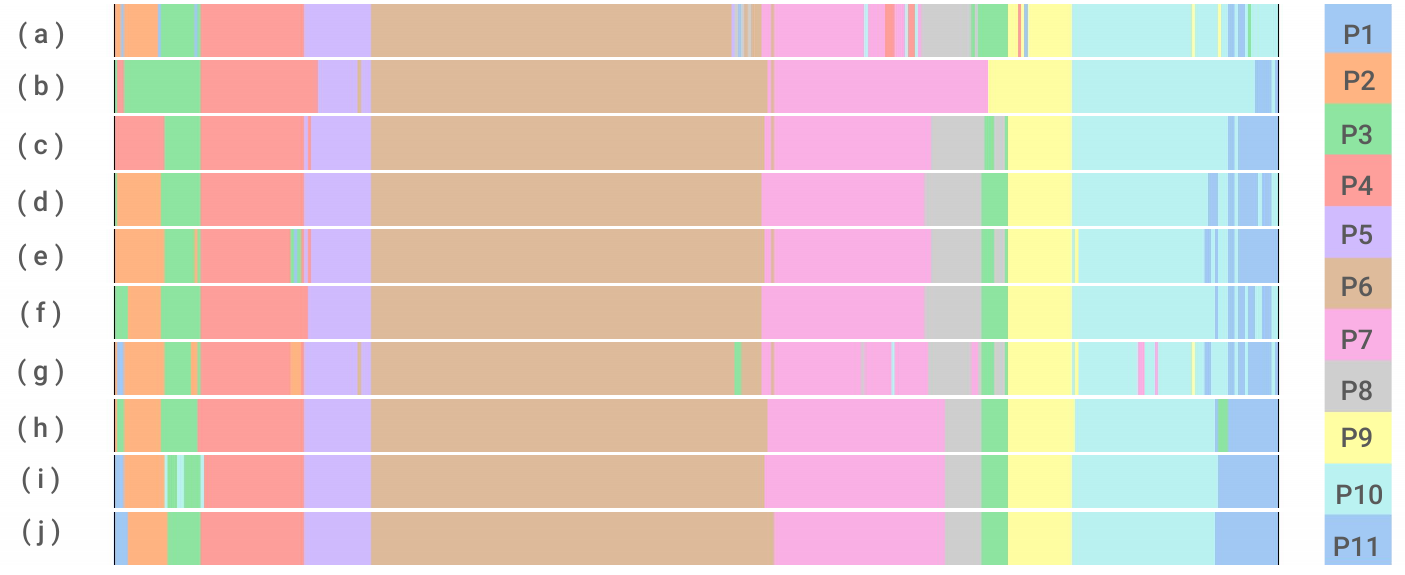}
 \vskip-10pt \caption{Qualitative results of step prediction of models on Cataracts-101 dataset in color-coded ribbon format. Predictions from (a) ResNet\cite{he2016deep}, (b) SV-RCNet\cite{jin2017sv}, (c) OHFM\cite{yi2019hard}, (d) TeCNO\cite{czempiel2020tecno}, (e) TMRNet\cite{jin2021temporal}, (f) Trans-SVNet\cite{gao2021trans}, (g) ViT\cite{dosovitskiy2020image}, (h) TimesFormer\cite{bertasius2021space}, (i) \GLSF, and (j) Ground Truth. P1 to P11 indicates Step label.}
  \label{fig:results}
\end{figure}
\noindent {\bf{Qualitative Comparison.}}
In Fig.~\ref{fig:results}, the color-coded ribbon map of one cataract surgery video from the cataract-101 dataset is shown. Both the ResNet and ViT produce noisy patterns and frequently generate abrupt predictions due to the lack of temporal relations. However, LSTM-based models such as SV-RCNet, OHFM, and TMRNet have comparatively less noisy patterns locally, but suffer from wrong step predictions due to the lack of complete global context. TeCNO and Trans-SVNet achieve smoother results by capturing long-term temporal information in later stages from the spatial embeddings
generated by the ResNet. However, these models still suffer from wrong step predictions (such as at P1) and noisy patterns (such as at P10) due to their high reliance on extracted spatial features and error propagation across stages in the model. Specifically, errors in the early stages of spatial modeling ResNet can propagate and accumulate across later stages, and lead to incorrect predictions. The ribbon plot of TimesFormer (i) demonstrates significant improvement compared to previous methods due to joint learning of spatio-temporal features. However, due to the lack of pivotal coarse long-term information, misclassifications are observed in local step transition areas, as seen in the incorrect prediction of P2 at locations of P1 and P11.  On the other hand, \GLSF elegantly aggregates spatio-temporal information from both streams, making the features more reliable. Additionally, our approach uses a single stage to limit the amount of error propagated across stages, contributing to improved accuracy in surgical step recognition.\\
\noindent {\bf{Ablation Studies.}} 
The top part of Table \ref{tab:ablation} shows the effect of different sampling rates in the long-term stream for step prediction in the Cataract-101 dataset. The results demonstrate that incorporating a coarse long-term stream is crucial for achieving significant performance gains compared to not using any long-term sequence (as in ViT and TimesFormer). Additionally, we observe that gradually increasing the sampling rate from 2 to 8 improves performance across all metrics, except for a slight decline at a sampling rate of 16. This decline may be due to a loss of information and noisy predictions resulting from the high number of frames skipped between each selected frame. Therefore, we chose a sampling rate of 8, as it provided the optimal balance between capturing valuable information and avoiding noise in our long-stream sequence.
\begin{table}[htp!]
\centering
\caption{Ablation testing results for different temporal gating mechanisms and long-term stream sampling rates on the Cataract-101 dataset.}
\resizebox{0.9\textwidth}{!}{%
\begin{tabular}{cccccccc} \toprule
\multicolumn{4}{c}{} & Accuracy & Precision & Recall & Jaccard \\ \midrule
\multirow{4}{*}{\rotatebox[origin=c]{0}{Sampling Rate}} & 2 & & & 91.16 $\pm$ 0.81 & 86.47 $\pm$ 0.85 & 87.25 $\pm$ 1.00 & 76.85 $\pm$ 1.33 \\
 & 4 & & & 91.43 $\pm$ 0.72 & 87.46 $\pm$ 0.91 & 87.20 $\pm$ 0.90 & 77.48 $\pm$ 1.26 \\
 & 8 & & & \textbf{92.91 $\pm$ 0.67} & \textbf{90.04 $\pm$ 0.71} & \textbf{89.45 $\pm$ 0.79} & \textbf{81.89 $\pm$ 0.92} \\
 & 16 & & & 91.35 $\pm$ 0.86 & 88.91 $\pm$ 0.92 & 88.27 $\pm$ 0.98 & 79.24 $\pm$ 1.10 \\ \midrule 
\multirow{3}{*}{\rotatebox[origin=c]{0}{Temporal Gating}} & Only short-term & & & 90.76 $\pm$ 1.05 & 85.38 $\pm$ 0.93 & 84.47 $\pm$ 0.95 & 75.97 $\pm$ 1.26 \\
 & No Gating & & & 90.24 $\pm$ 0.75 & 86.84  $\pm$ 0.88 & 84.39  $\pm$ 0.90 & 74.82  $\pm$ 1.13 \\
\multirow{2}{*}{\rotatebox[origin=c]{0}{Stream}}  & Fix Param. Gating & & & 91.52  $\pm$ 0.83 & 87.13   $\pm$ 0.81 & 88.41  $\pm$ 0.94 & 78.03   $\pm$ 1.04 \\
 & Feature Gating & & & \textbf{92.91 $\pm$ 0.67} & \textbf{90.04 $\pm$ 0.71} & \textbf{89.45 $\pm$ 0.79} & \textbf{81.89 $\pm$ 0.92} \\ \bottomrule  
\end{tabular}}
\label{tab:ablation}
\end{table}
\\
To evaluate our gating mechanism's effectiveness, we conducted ablation experiments with three different settings, as summarized in the bottom part of Table \ref{tab:ablation}. Initially, we passed both short-term and long-term stream features directly in the shared multi-head attention layer, but the model's performance was worse compared to the model trained only with short-term information (75.97 vs 74.82 Jaccard). The reason for this could be the lack of filtering to extract coarse temporal information. For instance, the long-term stream may contain noisy spatial information that is irrelevant to the temporal attention mechanism, which can affect the model's ability to attend to relevant information.
Additionally, incorporating a learnable gating parameter to regulate the flow of information between the short-term and long-term streams enhanced our model's performance by 4\%, enabling individual stream refinement through cross-stream information.
However, we observe that this approach has a limitation as the amount of cross-stream information sharing remains fixed during inference regardless of the quality of the feature representation in both streams at a particular time-frame. 
To address this limitation, we propose predicting gating parameters directly based on the spatio-temporal representation in both streams for that time frame. This approach allows for dynamic gating parameters, which means that at a particular time-point, short-term temporal feature representation can variably leverage the long-term coarse information as well can prioritize its own representation if it is more reliable. Improvement of 3\% Jaccard score is realized by using feature-based gating parameter estimation  in \GLSF compared to a fixed parameter gating mechanism. Our ablation study clearly highlights the significance of our gated temporal mechanism for feature refinement.

\section{Conclusion}
We propose \GLSF, a vision transformer-based method for recognizing surgical steps in complex videos. Our approach uses a gated temporal attention mechanism to integrate short and long-term cues, resulting in superior performance compared to recent LSTM and vision transformer-based approaches that only use short-term information. Our end-to-end joint learning captures spatial representations and sequential dynamics more effectively than multi-stage networks. We extensively evaluated \GLSF and found that it consistently outperformed state-of-the-art models for surgical step recognition.

\subsection*{Acknowledgements.}
This research was supported by a grant from the National Institutes of Health, USA; R01EY033065. The content is solely the responsibility of the authors and does not necessarily represent the official views of the National Institutes of Health.

\bibliographystyle{splncs04}
\bibliography{main}

\end{document}

%% file: commands.tex
\usepackage{color}
\usepackage[accsupp]{axessibility}  

\usepackage{microtype}
\usepackage{adjustbox}
\usepackage{listings}
\usepackage{pythonhighlight}
\usepackage{float}
\usepackage{tablefootnote}
\usepackage{tabularx}
\usepackage{booktabs}
\usepackage{graphicx}
\usepackage{float}
\usepackage{wrapfig}

\usepackage{graphicx}
\usepackage{amsmath}
\usepackage{amssymb}
\usepackage{booktabs}
\usepackage{textgreek}

\usepackage[capitalize]{cleveref}
\crefname{section}{Sec.}{Secs.}
\Crefname{section}{Section}{Sections}
\Crefname{table}{Table}{Tables}
\crefname{table}{Tab.}{Tabs.}

\newcommand{\GLSF}{$\mathsf{GLSFormer~}$}
\DeclareSymbolFont{Xlargesymbols}{OMX}{cmex}{m}{n}
\DeclareMathSymbol{\Xsum}{\mathop}{Xlargesymbols}{80}